\pdfoutput=1
\documentclass[conference,a4paper]{IEEEtran}
\IEEEoverridecommandlockouts

\usepackage{amsmath,amssymb,amsfonts}
\usepackage{algorithmic}
\usepackage{graphicx}
\usepackage{textcomp}
\usepackage{xcolor}
\usepackage{CJKutf8}
\usepackage{booktabs} 
\usepackage{array}    
\usepackage{booktabs} 
\usepackage{caption} 
\usepackage{booktabs} 

\begin{document}
\begin{CJK}{UTF8}{gbsn}

\title{From Metaphor to Mechanism: How LLMs Decode Traditional Chinese Medicine Symbolic Language for Modern Clinical Relevance}




\author{\IEEEauthorblockN{Jiacheng Tang}
\IEEEauthorblockA{\textit{School of Information Science and} \\
\textit{Engineering ,Shandong Normal University}\\
Jinan, China \\
jctang@stu.sdnu.edu.cn}
\and
\IEEEauthorblockN{Nankai Wu}
\IEEEauthorblockA{\textit{School of Pharmacy} \\
\textit{China Pharmaceutical University China}\\
Nanjing, China \\
email address or ORCID}
\and
\IEEEauthorblockN{Fan Gao}
\IEEEauthorblockA{\textit{Department of Civil Engineering} \\
\textit{The University of Tokyo Japan}\\
Tokyo, Japan \\
fangao0802@gmail.com}
\and
\IEEEauthorblockN{Chengxiao Dai}
\IEEEauthorblockA{\textit{Faculty of Information and Communication} \\
\textit{Technology, Universiti Tunku Abdul Rahman}\\
Kampar, Malaysia \\
daicx1226@1utar.my}
\and
\IEEEauthorblockN{Mengyao Zhao}
\IEEEauthorblockA{\textit{School of Information Science and} \\
\textit{Engineering ,Shandong Normal University}\\
Jinan, China \\
mengyaozhao@stu.sdnu.edu.cn}
\and
\IEEEauthorblockN{Xinjie Zhao\textsuperscript{*}}
\IEEEauthorblockA{\textit{Department of Civil Engineering} \\
\textit{The University of Tokyo Japan}\\
Tokyo, Japan \\
xinjie-zhao@g.ecc.u-tokyo.ac.jp}
}

\maketitle

\begin{abstract}
Metaphorical expressions are abundant in Traditional Chinese Medicine (TCM), conveying complex disease mechanisms and holistic health concepts through culturally rich and often abstract terminology. Bridging these metaphors to anatomically driven Western medical (WM) concepts poses significant challenges for both automated language processing and real-world clinical practice. To address this gap, we propose a novel multi-agent and chain-of-thought (CoT) framework designed to interpret TCM metaphors accurately and map them to WM pathophysiology. Specifically, our approach combines domain-specialized agents (TCM Expert, WM Expert) with a Coordinator Agent, leveraging stepwise chain-of-thought prompts to ensure transparent reasoning and conflict resolution. We detail a methodology for building a metaphor-rich TCM dataset, discuss strategies for effectively integrating multi-agent collaboration and CoT reasoning, and articulate the theoretical underpinnings that guide metaphor interpretation across distinct medical paradigms. We present a comprehensive system design and highlight both the potential benefits and limitations of our approach, while leaving placeholders for future experimental validation. Our work aims to support clinical decision-making, cross-system educational initiatives, and integrated healthcare research, ultimately offering a robust scaffold for reconciling TCM’s symbolic language with the mechanistic focus of Western medicine.
\end{abstract}


\section{Introduction}
Traditional Chinese Medicine (TCM) has, for centuries, offered a holistic paradigm for understanding human health, providing insights that extend beyond anatomical structures to encompass cultural, philosophical, and empirical observations \cite{matos2021understanding}. Central to TCM’s epistemology is its extensive reliance on metaphorical language, such as “liver fire flaring up” or “the Earth element failing to nourish the Metal.” Rather than simply ornamenting discourse, these metaphors serve as conceptual cornerstones that reflect a worldview shaped by Yin--Yang balance, Five Elements theory, and the dynamic principle of \textit{qi}. From this vantage point, systemic interactions and holistic harmony take precedence over the discrete, organ-focused models favored by Western medicine (WM).
Yet, TCM metaphors often defy direct anatomical or pathological translation, complicating cross-disciplinary communication. This challenge is exemplified by the divergent lenses through which TCM and WM interpret disease: the former employs descriptive, image-rich constructs to capture the body’s integrated functioning, while the latter adopts reductionist frameworks based on cellular and molecular pathophysiology \cite{guoan2011system}. As a result, TCM’s metaphor-driven descriptions risk being dismissed as vague or unscientific when viewed solely through the lens of WM’s quantifiable models. Such divergence has profound implications for clinical practice, education, and research, where practitioners and scholars frequently grapple with reconciling TCM’s culturally informed narratives and WM’s mechanistic rigor.

\begin{figure}[htbp]
\centering
\includegraphics[width=\columnwidth]{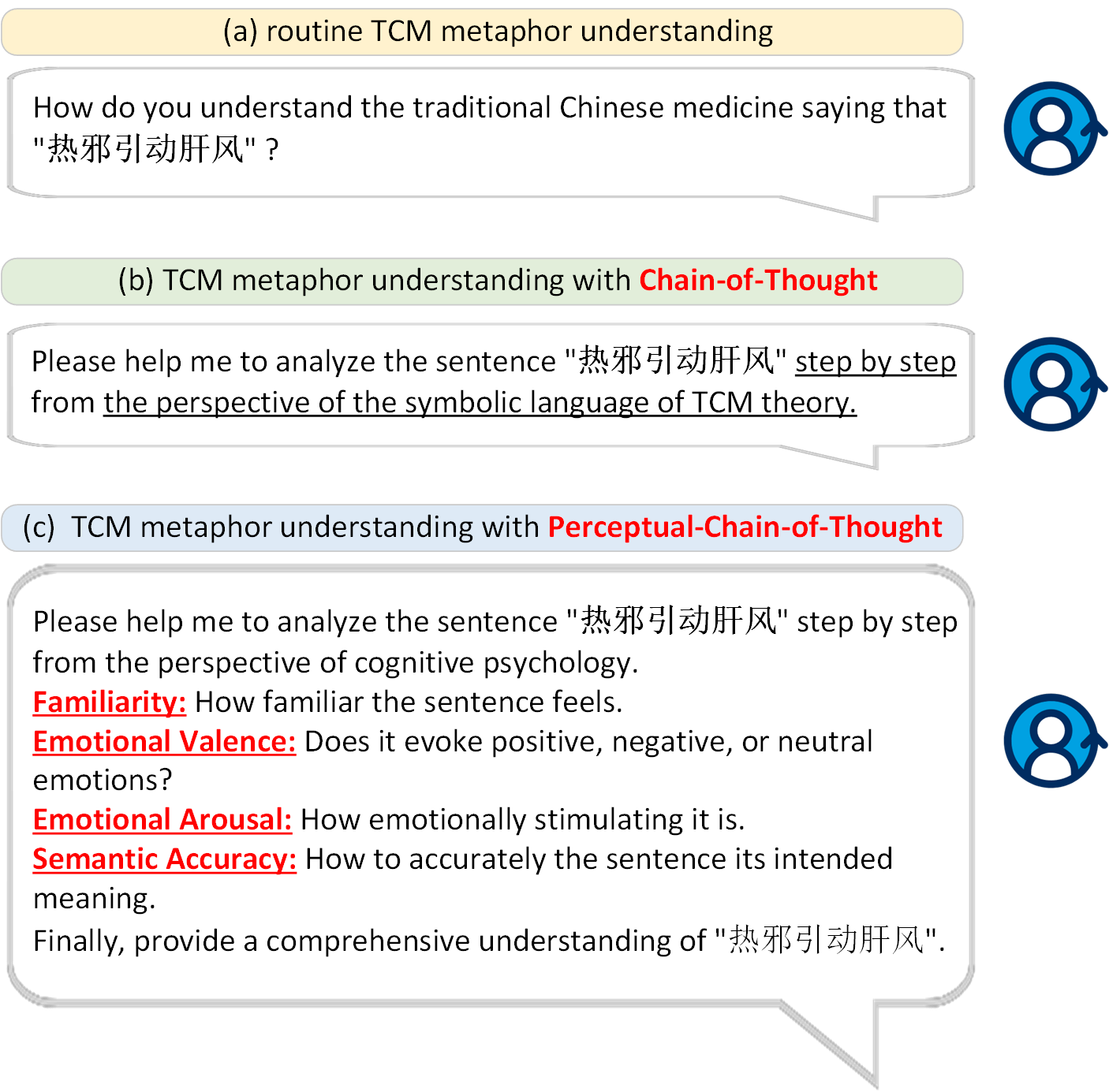}
\caption{Fig.(a) demonstrates the use of routine LLM methods to assist in understanding TCM metaphors. Fig.(b) demonstrates the use of COT to understand TCM metaphors. Fig.(c) demonstrates the Perceptual-COT to understand TCM metaphors}
\label{fig}
\end{figure}

\begin{figure*}[ht]
\centerline{\includegraphics[width=\textwidth]{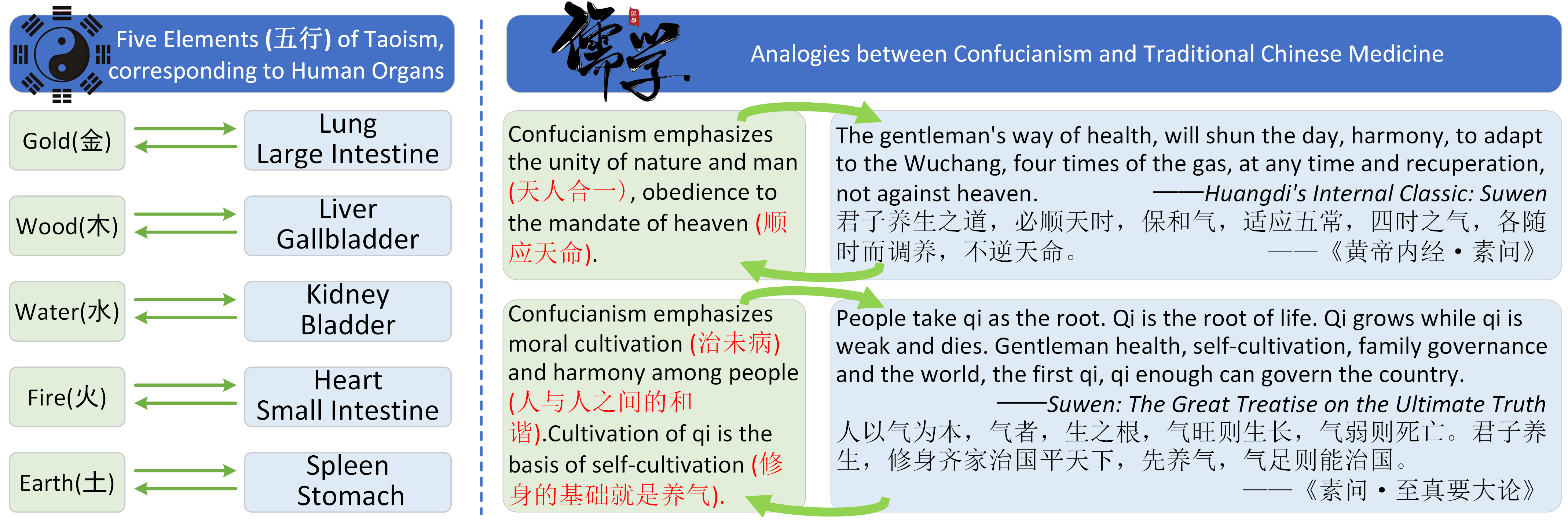}}
\caption{Comparison of metaphorical relationship between TCM and Taoism, TCM and Confucianism}
\label{fig}
\end{figure*}

Recent advancements in Natural Language Processing (NLP) and Large Language Models (LLMs) present a promising avenue for bridging these conceptual gaps \cite{haltaufderheide2024ethics}. While state-of-the-art LLMs have demonstrated exceptional proficiency in parsing and generating text grounded in Western biomedical nomenclature, these models typically fall short when interpreting metaphor-laden TCM descriptions. The complexity of TCM’s linguistic and philosophical heritage—manifested in classical Chinese texts, Confucian ethics, and Taoist notions of equilibrium—does not readily map onto the literal and data-driven corpora that train contemporary LLMs. Consequently, naive attempts to encode TCM metaphors risk generating misleading, superficial, or “hallucinated” mappings that fail to convey deeper TCM principles.
Addressing this fundamental challenge demands a methodological synthesis of advanced AI techniques and TCM theory. Within language philosophy, metaphors are not mere stylistic devices but serve as cognitive tools that reveal how cultures conceptualize abstract phenomena \cite{lakoff1980metaphors}. In the TCM domain, these conceptual layers encompass notions of interdependency, balance, and systemic flow, all of which are difficult to capture via purely literal or reductionist modeling. Thus, an integrative approach must harness LLMs’ capabilities—such as long-range contextual reasoning and compositional inference—while preserving the rich philosophical underpinnings that give TCM metaphors their explanatory power.


This paper introduces a novel multi-agent and chain-of-thought (CoT) framework aimed at reconciling these challenges. Within our approach, a dedicated \emph{TCM Expert} agent interprets metaphorical expressions in light of TCM’s classical doctrines, while a \emph{WM Expert} agent grounds these interpretations within contemporary biomedical science. A \emph{Coordinator} agent then synthesizes the two outputs, revealing conceptual alignments or discrepancies. Central to this design is the explicit recording of each agent’s reasoning steps, a CoT mechanism that promotes transparency and robust cross-verification. The resulting system mitigates hallucination errors through iterative refinement and domain-informed logic checks, enabling more faithful translations between TCM’s metaphor-rich discourse and WM’s mechanistic categories.

\section{Related Work}

Research in metaphorical reasoning and large language models (LLMs) has evolved significantly over the past decade, yet most endeavors center on Western languages, rhetorical structures, and biomedical paradigms. In contrast, Traditional Chinese Medicine (TCM) employs metaphor not merely as a figurative device but as a foundational tool to conceptualize a spectrum of syndromes and therapies. This section synthesizes relevant literature into two primary themes: (1) how TCM metaphors have been studied and computationally modeled, and (2) how LLMs and multi-agent systems have been applied in medical contexts, with a focus on bridging TCM and Western Medicine (WM).

\subsection{Metaphorical Reasoning in TCM and Beyond}
Metaphors in TCM stand at the intersection of philosophy, clinical practice, and cultural symbolism. They serve to translate intricate ideas—ranging from Yin--Yang and Five Elements to meridian pathways—into linguistically accessible yet conceptually layered constructs \cite{mykhalchuk2021cross, su2019chinese, gao2020recent}. For instance, terms such as “liver wind stirring internally” or “fire blazing in the heart” transcend literal organ references, invoking holistic paradigms where bodily functions, emotional states, and environmental factors converge \cite{feng2019ontology, nichols2021understanding}. 

In computational research, metaphor detection in English has generally focused on syntactic and semantic markers \cite{wills2010information, tang2024medagentslargelanguagemodels, le2020multi}, supplemented by external knowledge bases for domain-specific interpretation \cite{cotter2021framing, li2017demystifying}. However, standard approaches often fail to capture the brevity, classical Chinese syntax, and multi-layered cultural context that shape TCM metaphors \cite{xue2019fine, chen2018evaluating}. Early attempts to incorporate TCM into computational frameworks—such as building structured ontologies or knowledge graphs—have yielded partial solutions but often overlook metaphorical richness \cite{feng2019ontology, liu2024research}. More recent studies leverage advanced word embeddings fine-tuned on historical TCM texts, revealing improved robustness in capturing subtle conceptual cues \cite{chu2020quantitative, banerjee2025pharmaceutical}. Nonetheless, widespread challenges persist, including scant annotated corpora, polysemous terminology, and heterogeneous TCM doctrines \cite{zeilig2022dementia, xu2022epistemic}.
Methodologically, researchers are exploring topic modeling, latent semantic analysis, and cross-lingual transfer learning to adapt Western-centric metaphor processing algorithms to TCM corpora \cite{mm2023effects, liao2023comparative}. While these techniques have demonstrated promise—for instance, improved detection of figurative expressions in Chinese medical documents—they remain limited in their ability to handle the profound cultural embedding of TCM metaphors \cite{tsui2024discourse, zhang2022information}. Consequently, new strategies are needed to systematically address TCM’s multilayered conceptual scheme, particularly in bridging abstract notions (e.g., Yin deficiency) with empirical or mechanistic frameworks \cite{bao2023intelligent, wang2023novel}.

\subsection{LLMs in Medical Domains}
With the advent of large-scale pretrained language models, medical NLP has rapidly evolved to include clinical text summarization, automated diagnosis, and specialized question-answering systems \cite{yuan2025natural, wang2024hct}. Groundbreaking architectures such as GPT-style models and BERT derivatives achieve high performance in English-based tasks like entity recognition and relation extraction \cite{xie2024organic, jin2024better}. However, these models generally rely on standardized terminologies (e.g., ICD codes, SNOMED CT) and extensive biomedical corpora, which are sparse in TCM settings \cite{zhang2023comparative, xu2022epistemic}. As a result, LLMs pretrained on predominantly Western medical literature risk generating “hallucinations” when confronted with culturally nuanced, metaphor-heavy TCM input \cite{fisher2019explaining, bao2023intelligent}.
In an effort to align TCM concepts with WM data, researchers have experimented with bilingual corpora, co-training paradigms, and cross-lingual embeddings \cite{cotter2021framing, mykhalchuk2021cross}. While these approaches improve lexical alignment, they often fail to preserve the symbolic depth of TCM. Embedding-based models, for example, may learn semantic proximities between TCM entities and Western pathophysiological terms, but they struggle to represent the underlying philosophical constructs \cite{nichols2021understanding, chen2018evaluating}. Moreover, direct one-to-one mappings (e.g., equating “Spleen Qi deficiency” to “digestive malfunction”) risk oversimplifying the multifaceted role of Spleen in TCM theory \cite{su2019chinese, le2020multi}.

\subsubsection{Multi-Agent Systems and Chain-of-Thought Paradigms}
A promising development in addressing TCM’s metaphorical complexity within medical NLP is the rise of multi-agent systems \cite{tang2024medagentslargelanguagemodels, liao2023comparative}. Such frameworks delegate distinct tasks—like metaphor parsing, biomedical validation, and integrative reasoning—to specialized agents, each fine-tuned or trained on relevant corpora \cite{wills2010information, liu2024research}. By sharing intermediate inferences, these agents collectively reduce misinterpretations that often occur when a single model attempts to span diverse linguistic and conceptual domains \cite{gao2020recent, bao2023intelligent}.
Chain-of-thought (CoT) prompting further augments multi-agent architectures by making the inference process transparent \cite{feng2019ontology, mykhalchuk2021cross}. This is especially pertinent for TCM, where diagnostic interpretations are rarely linear. Metaphors like “water failing to moisten wood” may imply internal dryness, mental unrest, or both, depending on the broader clinical context \cite{li2017demystifying, banerjee2025pharmaceutical}. By explicitly enumerating reasoning steps, CoT prompting allows for cross-verification and iterative refinement, mitigating the risk of oversimplification or context collapse \cite{jin2024better, su2019chinese}.

\begin{figure*}[htbp]
\centerline{\includegraphics[width=0.9\textwidth]{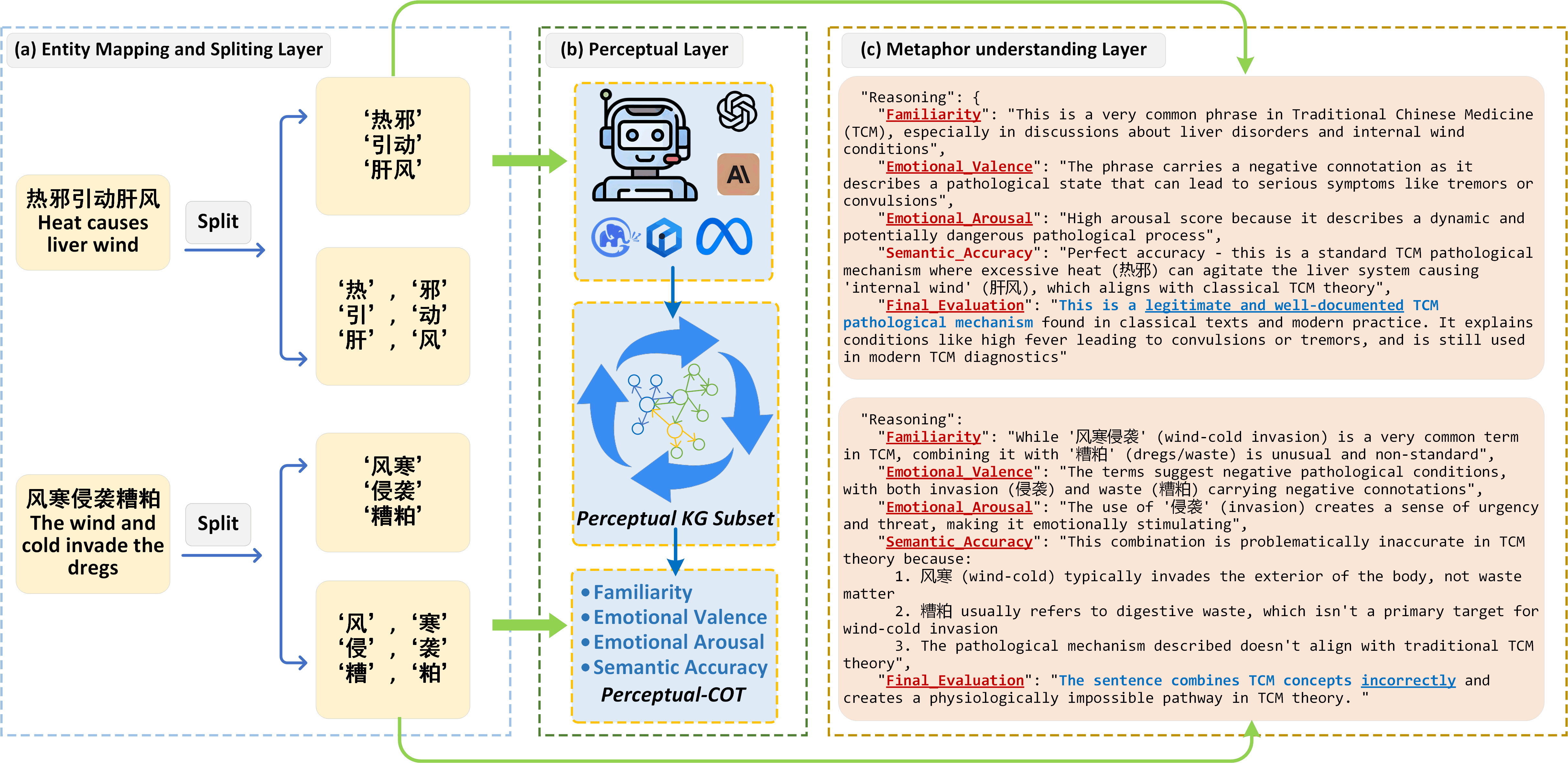}}
\caption{Our proposed Perceptual-Chain-Of-Thought framework}
\label{fig}
\end{figure*}

\section{Methodology}

In this section, we present a unified, math-driven methodology that formalizes TCM concepts and reasoning processes, organizes them into a coherent multi-agent architecture, and incorporates a chain-of-thought (CoT) mechanism to enhance transparency and interpretability. By grounding TCM’s metaphorical and philosophical underpinnings in rigorous semantics, symbolic logic, and probabilistic or fuzzy inference, this framework provides an integrative bridge between TCM and Western medicine (WM).

\subsection{Multi-Dimensional Semantic Representation of TCM Concepts}
Let $ \mathcal{C} = \{c_1, c_2, \dots, c_m\} $ denote the set of TCM concepts, including syndromes, meridians, pathogenic factors, and key theoretical notions. Each concept $ c_i $ is associated with a high-dimensional vector $ \mathbf{v}_i \in \mathbb{R}^d $, constructed via domain-adapted embedding methods (e.g., BERT, ERNIE, or knowledge graph embeddings) combined with curated TCM corpora. Formally, we define an embedding function
$$
\Phi: \mathcal{C} \rightarrow \mathcal{V} \subset \mathbb{R}^d,
$$
where $ \Phi(c_i) = \mathbf{v}_i $ characterizes the semantic relationship among TCM entities based on their contextual similarities and symbolic connections. To capture polysemy and cultural nuances, we augment these embeddings with specialized tokens for concepts such as “yin deficiency” or “liver fire,” and use classical Chinese references to ensure authenticity.

Additionally, certain TCM concepts map to discrete symbolic labels. Let $ \Sigma = \{\sigma_1, \sigma_2, \dots, \sigma_k\} $ represent a finite set of symbolic descriptors (e.g., “excess,” “deficiency,” “hot,” “cold,” “internal,” “external”). Each $ c_i $ may be assigned one or more labels from $ \Sigma $ to reflect classical TCM categorization. Hence, a hybrid semantic representation is formed by combining continuous vectors $ \mathbf{v}_i $ with symbolic tags $ \{\sigma_j\} $.

To capture TCM’s complex diagnostic and therapeutic logic—where relationships are often uncertain, context-dependent, and gradational—we introduce a layered framework encompassing symbolic logic, fuzzy sets, and Bayesian networks.

\subsubsection{Symbolic Logic Layer}

We denote atomic propositions $ p_i $ to represent elementary TCM statements such as “patient has damp-heat in the liver” or “tongue coating is yellow and greasy.” Using standard logical connectives $ (\land, \lor, \neg, \implies) $, we can construct composite statements or rules that reflect classical TCM diagnostic schemas, for example:
$$
p_{\text{damp-heat}} \land p_{\text{tongue}} \implies p_{\text{liver-syndrome}}.
$$
These rules may be derived from canonical texts or expert knowledge. The symbolic logic layer thus encodes deterministic or rule-based relationships widely accepted within TCM scholarship.

\subsubsection{Fuzzy Logic and Uncertainty Representation}

Since TCM diagnoses often hinge on qualitative judgments and partial truths (e.g., mild dampness, moderate heat), we introduce fuzzy sets. For each proposition $ p_i $ and evidence set $ E $, we define a membership function
$$
\mu(p_i, E): \mathcal{C} \times \mathcal{E} \rightarrow [0,1],
$$
where $ \mu(p_i, E) $ quantifies how strongly the evidence $ E $ supports the proposition $ p_i $. This function might map clinical signs like pulse quality or tongue color to a numeric scale, aligned with TCM’s notion of “excess” or “deficiency.” Logical inference then proceeds via fuzzy operations:
$$
\mu(p_i \land p_j, E) = \min \bigl(\mu(p_i, E), \mu(p_j, E)\bigr),
$$
$$
\mu(p_i \lor p_j, E) = \max \bigl(\mu(p_i, E), \mu(p_j, E)\bigr),
$$
subject to modifications depending on specific TCM contexts. Thus, fuzzy logic respects the gradational and often imprecise nature of TCM syndromes.

\subsubsection{Bayesian Network Layer}

To incorporate probabilistic reasoning and causality, particularly in scenarios that blend TCM and WM evidence, we introduce a Bayesian network with node set $ \mathcal{N} $ and directed edges indicating causal or correlative influences. Each node $ n \in \mathcal{N} $ may represent a TCM syndrome, a WM lab finding, or a bridging concept (e.g., “inflammatory marker elevated due to damp-heat”). A conditional probability distribution
$$
P(n_j \mid n_i)
$$
encodes how node $ n_i $ affects node $ n_j $ through either TCM rationale (e.g., “heat in the stomach can cause vomiting”) or known biomedical pathways (e.g., “inflammatory cytokine levels affect body temperature”). We update these distributions via Bayesian inference:
$$
P(n_j \mid \text{evidence}) \propto P(\text{evidence} \mid n_j) \, P(n_j),
$$
and iteratively integrate TCM and WM data within a unified probabilistic graph. This structure allows flexible incorporation of lab tests, imaging findings, or TCM pulse diagnoses, thus offering a robust framework for cross-system reasoning.

To ensure consistency, we represent the entire knowledge space by a meta-language $ \mathcal{M} $ that integrates semantic embeddings, fuzzy membership functions, and Bayesian conditional distributions. Formally, $ \mathcal{M} $ can be characterized as a tuple
$$
\mathcal{M} = \bigl(\mathcal{C}, \mathcal{V}, \Sigma, \mathcal{P}, \mu, P, \mathcal{R}\bigr),
$$
where $ \mathcal{C} $ is the set of TCM concepts, $ \mathcal{V} \subset \mathbb{R}^d $ is the embedding space, $ \Sigma $ is the set of symbolic labels, $ \mathcal{P} $ is the set of propositions, $ \mu $ is the fuzzy membership function, $ P $ denotes Bayesian relationships, and $ \mathcal{R} $ includes logical inference rules. All modules operate under these shared definitions to avoid contradictory interpretations or repeated definitions. When novel syndromes, herbs, or experimental data arise, they can be seamlessly added to $ \mathcal{M} $, given the appropriate embeddings $ \mathbf{v}_i $, membership calibrations $ \mu $, and conditional probability updates $ P(\cdot \mid \cdot) $.

\subsection{Multi-Agent Implementation Strategy}

Although the core of this methodology lies in mathematical and logical representations of TCM, the practical realization of these representations is facilitated by a multi-agent ecosystem. We define a set of agents $ \mathcal{A} = \{A_1, A_2, \dots, A_m\} $, each responsible for a different segment of the TCM cognitive process and integrated under a shared knowledge base grounded in $ \mathcal{M} $. For instance, a Knowledge-Extraction Agent parses classical texts and clinical notes to populate $ \mathbf{v}_i $ embeddings and update fuzzy membership functions $ \mu $. A Reasoning Agent focuses on applying logical rules or Bayesian updates to arrive at potential diagnoses or treatments, while an Evaluation Agent measures model performance, tracks metrics such as diagnostic accuracy, and ensures the consistency of fuzzy and probabilistic outputs. A Coordinator Agent orchestrates inter-agent communication, synchronizing the chain-of-thought logs and reconciling TCM inferences with WM interpretations.

\subsection{Chain-of-Thought (CoT) Integration}

Chain-of-thought prompting enriches the multi-agent system by transparently exposing intermediate reasoning. Each agent records a stepwise rationale for inferences made, including fuzzy membership calculations or updates to conditional probabilities. When the Reasoning Agent posits that $ p_{\text{liver-fire}} $ is true to a high degree (e.g., $ \mu(p_{\text{liver-fire}}, E) = 0.8 $), it stores a short textual or symbolic explanation: “Patient exhibits irritability, bitter taste, red tongue tip; these imply $ \text{liver-fire} $.” The WM Specialist Agent then attempts to map “liver-fire” to a plausible Western correlate, e.g., “inflammatory hepatic response,” detailing its own inference steps through $ P(n_{\text{inflammation}} \mid n_{\text{irritability}}, n_{\text{lab-values}}) $. The Coordinator Agent merges these chain-of-thought logs and highlights both convergences (e.g., systemic heat/inflammation) and discrepancies (e.g., no direct WM lab indicator for “fire”). By preserving explicit intermediate steps, CoT prompts allow domain experts to validate or revise the logical transitions, thus minimizing the risk of hidden “black box” reasoning.

\begin{table*}[htbp]
  \centering
  \caption{The constructed metaphor dataset}
    \begin{tabular}{p{11.875em} p{14.125em} p{11.75em} p{16.625em}}
    \toprule
    \multicolumn{2}{c}{\textbf{TCM terms}} & \multicolumn{2}{c}{\textbf{WM terms}} \\
    \cmidrule(lr){1-2} \cmidrule(lr){3-4}
    \textbf{Chinese term expression} & \textbf{English term expression} & \textbf{Chinese term expression} & \textbf{English term expression} \\
    \midrule
    脾脏运化水谷 & Spleen transport water valley & 脑室分布脑脊液 & Ventricles distribute cerebrospinal fluid \\
    甘草调和药性 & Licorice blends herbs & 脊髓传递神经信号 & Spinal cord transmits nerve signals \\
    滑石降土安气 & Talc reduces soil and air & 抗体中和病原体 & Antibody neutralizing pathogen \\
    金性沉降下行 & Gold settles down & 钙增强骨骼硬度 & Calcium strengthens bones \\
    寒湿凝结坚 & Cold and damp condense hard & 肝脏分泌胆汁 & Bile secretion by liver \\
    阳气迅速退敛 & Yang quickly recedes & 肾脏吸收氨基酸 & Renal absorption of amino acids \\
    风湿寒热邪 & Rheumatism cold heat evil & 维生素A增强免疫力 & Vitamin A boosts immunity \\
    血流营气赖 & Blood flow is full of qi & 尿酸排出代谢物 & Uric acid gets rid of metabolites \\
    邪气盘结成癥 & Evil qi disc form disease & 胃酸溶解细菌 & Stomach acid dissolves bacteria \\
    膀胱藏津液 & The bladder contains body fluid & 多巴胺调节情绪行为 & Dopamine regulates emotional behavior \\
    \bottomrule
    \end{tabular}%
  \label{tab:addlabel}%
\end{table*}%

\subsection{Experimental Framework}

We validate this methodology using a multimodal dataset $ \mathcal{D} $ collected from classical TCM treatises (e.g., \textit{Huangdi Neijing}, \textit{Shanghan Lun}), modern clinical records, prescription databases, and parallel WM health data. After hierarchical annotation, $ \mathcal{D} $ is split into training $ \mathcal{D}_{\text{train}} $ and testing $ \mathcal{D}_{\text{test}} $ sets using stratified sampling to capture diverse syndromes and prescriptions. Training involves jointly optimizing embeddings $ \mathbf{v}_i $ (via gradient-based methods), adjusting fuzzy membership functions $ \mu $ (through a combination of expert calibration and data-driven tuning), and refining Bayesian network parameters $ P(\cdot \mid \cdot) $ (using maximum likelihood or EM algorithms). We compare performance against baseline models that use either purely symbolic rules or purely deep learning networks without TCM context. Evaluation metrics include diagnostic accuracy, interpretive clarity (quantified by human experts reading chain-of-thought outputs), and the robustness of cross-system mappings.

Throughout the methodological pipeline, we maintain consistency by disallowing contradictory definitions in $ \mathcal{M} $ and checking the logical coherence of added or updated inference rules. When new syndromes emerge (e.g., modern reinterpretations of “COVID-19 in TCM contexts”), we incorporate them by assigning vectors $ \mathbf{v}_{\text{new}} $ through domain-adapted embeddings, calibrating fuzzy membership degrees $ \mu $ based on emerging clinical data, and linking them into the Bayesian network with conditional probabilities $ P(\text{new-node} \mid \text{existing-nodes}) $. This modular extensibility ensures that the framework is not only theoretically sound but also practically adaptable to evolving TCM scholarship and integrative medical research.

\begin{table*}[htbp]
  \centering
  \caption{Results on the TCM dataset with Perceptual-COT}
    \begin{tabular}{cccccccc}
    \toprule
    \textbf{LLM\_Models} & \textbf{Familiarity} & \textbf{Emotional Valence} & \textbf{Emotional Arousal} & \textbf{Semantic Accuracy} & \textbf{Accuracy(\%)} & \textbf{Recall(\%)} & \textbf{F1-score(\%)} \\
    \midrule
    ChatGLM3-6B & 6.77  & 1.64  & 1.78  & 7.61  & 55.23  & 18.76  & 29.70  \\
    DeepSeek-V3 & 7.84  & 5.39  & 6.11  & 8.23  & 79.79  & 77.58  & 79.34  \\
    HuatuoGPT2-7B & 7.36  & 5.39  & 5.20  & 7.92  & 50.17  & 68.75  & 59.78  \\
    LLaMA3.1-8B & 6.55  & 3.10  & 3.33  & 8.59  & 63.56  & 86.17  & 70.28  \\
    Mistral-7B & 3.62  & 3.37  & 4.13  & 3.72  & 56.02  & 9.40  & 17.18  \\
    Baichuan2-13B-Chat & 7.44  & 5.44  & 5.39  & 8.46  & 54.10  & 76.00  & 65.42  \\
    Qwen-7B-Chat & 3.92  & 3.54  & 2.29  & 7.86  & 82.41  & 81.25  & 81.68  \\
    ChatGPT-4o & 6.34  & 4.31  & 3.72  & 7.11  & 69.25  & 49.09  & 61.60  \\
    \bottomrule
    \end{tabular}%
  \label{tab:addlabel}%
\end{table*}%

\begin{table}[htbp]
  \centering
  \caption{Results on the TCM dataset without COT}
    \begin{tabular}{cccc}
    \toprule
    \textbf{LLM\_Models} & \textbf{Accuracy(\%)} & \textbf{Recall(\%)} & \textbf{F1-score(\%)} \\
    \midrule
    ChatGLM3-6B & 52.74  & 15.07  & 24.14  \\
    DeepSeek-V3 & 76.84  & 74.63  & 76.32  \\
    HuatuoGPT2-7B & 43.43  & 62.50  & 54.35  \\
    LLaMA3.1-8B & 61.79  & 84.40  & 68.84  \\
    Mistral-7B & 54.20  & 7.52  & 13.75  \\
    Baichuan2-13B-Chat & 50.29  & 72.67  & 62.55  \\
    Qwen-7B-Chat & 72.36  & 70.83  & 71.20  \\
    ChatGPT-4o & 66.21  & 46.06  & 57.79  \\
    \bottomrule
    \end{tabular}%
  \label{tab:addlabel}%
\end{table}%

\begin{table*}[htbp]
  \centering
  \caption{Results on the WM dataset with Perceptual-COT}
    \begin{tabular}{cccccccc}
    \toprule
    \textbf{LLM\_Models} & \textbf{Familiarity} & \textbf{Emotional Valence} & \textbf{Emotional Arousal} & \textbf{Semantic Accuracy} & \textbf{Accuracy(\%)} & \textbf{Recall(\%)} & \textbf{F1-score(\%)} \\
    \midrule
    ChatGLM3-6B & 7.08  & 0.80  & 1.36  & 8.13  & 61.72  & 35.31  & 48.41  \\
    DeepSeek-V3 & 6.35  & 4.88  & 3.19  & 7.58  & 98.23  & 96.46  & 98.20  \\
    HuatuoGPT2-7B & 7.39  & 5.28  & 5.16  & 4.56  & 50.07  & 49.75  & 51.22  \\
    LLaMA3.1-8B & 6.88  & 1.72  & 3.36  & 8.73  & 90.68  & 85.35  & 90.25  \\
    Mistral-7B & 5.90  & 1.84  & 3.92  & 6.97  & 72.29  & 44.05  & 61.00  \\
    Baichuan2-13B-Chat & 7.50  & 4.58  & 4.81  & 8.57  & 54.76  & 58.85  & 57.20  \\
    Qwen-7B-Chat & 4.79  & 3.04  & 2.03  & 7.86  & 75.90  & 52.10  & 68.51  \\
    ChatGPT-4o & 7.41  & 3.56  & 3.53  & 6.83  & 95.12  & 92.28  & 94.96  \\
    \bottomrule
    \end{tabular}%
  \label{tab:addlabel}%
\end{table*}%

\begin{table}[htbp]
  \centering
  \caption{Results on the WM dataset without COT}
    \begin{tabular}{cccc}
    \toprule
    \textbf{LLM\_Models} & \textbf{Accuracy(\%)} & \textbf{Recall(\%)} & \textbf{F1-score(\%)} \\
    \midrule
    ChatGLM3-6B & 53.22  & 16.77  & 26.54  \\
    DeepSeek-V3 & 76.84  & 74.63  & 76.32  \\
    HuatuoGPT2-7B & 43.43  & 62.50  & 54.35  \\
    LLaMA3.1-8B & 61.79  & 84.40  & 68.84  \\
    Mistral-7B & 70.37  & 42.11  & 58.30  \\
    Baichuan2-13B-Chat & 50.29  & 72.67  & 62.55  \\
    Qwen-7B-Chat & 72.36  & 70.83  & 71.20  \\
    ChatGPT-4o & 92.16  & 89.32  & 91.91  \\
    \bottomrule
    \end{tabular}%
  \label{tab:addlabel}%
\end{table}%

\section{Experiments}
\subsection{Experimental Setup}

\subsubsection{Dataset}
To validate our unified, mathematically driven approach for integrating Traditional Chinese Medicine (TCM) and Western Medicine (WM), we constructed a dataset of 2,801 sentences. These sentences were meticulously selected and categorized to capture distinct theoretical alignments and varying degrees of linguistic complexity.

\textbf{Data Sources:} TCM terminologies were extracted from authoritative textbooks and classical works published by the China Traditional Chinese Medicine Press, including \textit{Basic Theory of Traditional Chinese Medicine} (7th Edition), \textit{Formulas of Traditional Chinese Medicine}, \textit{Chinese Materia Medica}, \textit{Diagnostics of Traditional Chinese Medicine} (7th Edition), \textit{Internal Medicine of Traditional Chinese Medicine} (7th Edition), and classical texts such as \textit{Huangdi Neijing}, \textit{Shanghan Lun}, \textit{Jinkui Yaolue}, \textit{Wenbing Lun}, and \textit{Wenbing Tiaobian}, as well as the \textit{Dictionary of Traditional Chinese Medicine}. WM terminologies were derived from the \textit{English-Chinese Medical Dictionary} (published by the People’s Medical Publishing House) and from several medical textbooks by the same publisher, including \textit{Pathology} (7th Edition), \textit{Internal Medicine} (7th Edition), and \textit{Diagnostics} (7th Edition).

\textbf{Selection Criteria:} Candidate terminologies were first filtered based on a subject-verb-object (SVO) structure. Only terms pertinent to disease etiology, pathogenesis, pathology, pathophysiology, and clinical manifestations were retained. This ensures that the sentences reflect both TCM and WM conceptualizations of illness. We adopted a four-stage layered screening. Initially, candidate terms were collected from professional dictionaries, textbooks, and domain-specific literature. Next, medical experts performed a preliminary filter to preserve terms predominantly used in medicine, removing those deemed semantically redundant. From an initial pool of 3,000 terms, 199 were eliminated due to duplication or irrelevance. Finally, domain specialists verified the remaining terms, resulting in a balanced dataset of 613 correct TCM terms, 703 incorrect TCM terms, 697 correct WM terms, and 788 incorrect WM terms.

The final set of 2,801 sentences, evenly distributed across four categories—sentences aligning with TCM theory, sentences misaligned with TCM, sentences aligning with WM theory, and sentences misaligned with WM—was derived from the curated terminologies. Each sentence was formulated using a consistent SVO structure (e.g., “The spleen and stomach transport water and grain”), thereby reducing syntactic variability and improving clarity for both manual and automated analyses.

\subsubsection{Baselines}
To comprehensively assess how contemporary large language models interpret TCM metaphors, we selected six models covering different parameter sizes and application scenarios:
\textbf{Baichuan2-7B-Chat}, \textbf{Qwen-7B-Chat}, \textbf{ChatGLM3-6B}, \textbf{HuatuoGPT2-7B}, \textbf{Llama3.1-8B}, \textbf{Mistral-7B}.

\subsubsection{Metrics}
We employ three primary metrics to assess model performance in interpreting TCM metaphors: \textit{Accuracy}, \textit{Recall}, and the \textit{F1} score.

\begin{enumerate}
    \item \textbf{Accuracy:} Proportion of correctly predicted samples among all samples:
    \[
    \text{Accuracy} = \frac{\text{TP} + \text{TN}}{\text{TP} + \text{FP} + \text{FN} + \text{TN}}
    \]
    \item \textbf{Recall:} Fraction of actual positive samples (i.e., correct metaphors) that are correctly identified:
    \[
    \text{Recall} = \frac{\text{TP}}{\text{TP} + \text{FN}}
    \]
    A higher Recall indicates the model’s ability to detect a broader range of valid TCM metaphors.

    \item \textbf{F1 Score:} The harmonic mean of Precision and Recall:
    \[
    \text{F1} = 2 \times \frac{(\text{Precision}) \times (\text{Recall})}{(\text{Precision}) + (\text{Recall})}
    \]
    Although Precision is not explicitly shown, the F1 score offers a balanced measure between how many predictions are correct versus how many correct instances are captured, thereby providing a robust assessment of overall model efficacy.
\end{enumerate}

\subsubsection{Implementation Details}

To address the observed limitations of current LLMs in handling metaphor-rich TCM discourse, we fine-tuned Baichuan2-7B-Chat on a dedicated \textit{metaphor comprehension} module and a \textit{semantic relevance} module. We employed LoRA-based techniques, quantized to 8 bits, with low-rank adaptation (rank=12, alpha=24). The learning rate was set to 2e-4, batch size to 64, and maximum sequence length to 1,024. Our training strategy involved an initial mixed phase of one epoch, followed by two epochs each dedicated to metaphor understanding and semantic relevance tasks. Implementations relied on Transformers, PEFT, and DeepSpeed frameworks, running on eight 4090 GPUs (24 GB) in parallel. During semantic relevance evaluation, Top-k was set to 5. This configuration ensures that the model effectively learns to capture metaphorical nuances in TCM while preserving computational feasibility.

\subsection{Ablation Study}
The next research phase involves a set of both quantitative and qualitative evaluations of our multi-agent architecture.  We divide the metaphor-rich dataset into training, validation, and test sets, ensuring an equitable representation of diverse TCM syndromes and WM conditions.
\textbf{Model Implementation:} We apply either fine-tuning or prompt-engineering approaches on large-scale LLMs (e.g., GPT-4, BERT variants) for each specialized agent. Varying corpus weights for TCM and WM materials will also be explored.
\textbf{Evaluation Metrics:} 
    \begin{itemize}
        \item \textit{Metaphor Parsing Accuracy:} The system’s ability to detect and interpret TCM metaphors with fidelity.
        \item \textit{Cross-System Diagnostic Correctness:} Accuracy of WM-based interpretations, measured against clinically validated ground truths.
        \item \textit{Human Expert Judgments:} TCM and WM professionals will appraise the coherence and plausibility of integrative outputs.
    \end{itemize}
\textbf{Baseline Comparisons:} Single-agent or direct LLM solutions without multi-agent collaboration or chain-of-thought (CoT) prompts, used to gauge the effectiveness of our structured framework.

\section{Results and Analysis}

Table II and Table III summarize the performance of eight models on the TCM dataset, comparing results obtained with and without chain-of-thought (CoT) prompts, while Table IV and Table V provide analogous results for the Western Medicine (WM) dataset. In both TCM and WM contexts, most models show significant improvements in Accuracy, Recall, and F1 scores when CoT is employed. For instance, on the TCM dataset, Qwen-7B-Chat achieves an Accuracy of 82.41\% and F1 of 81.68\% with CoT, compared to 72.36\% and 71.20\% without CoT. A similar pattern emerges in the WM dataset for ChatGPT-4o, whose Accuracy increases from 92.16\% to 95.12\% once CoT reasoning is introduced. Even more strikingly, DeepSeek-V3 exhibits a jump from 76.84\% to 98.23\% in Accuracy on the WM dataset under CoT conditions, underscoring the benefits of providing intermediate reasoning steps in refining medical inferences.
Although the overall trend favors CoT-enhanced models, there are notable variations. On the TCM dataset, Mistral-7B and HuatuoGPT2-7B struggle to maintain high Recall, indicating that certain metaphorical expressions or domain-specific idioms remain challenging for smaller or less domain-aligned models. For example, Mistral-7B achieves only a 9.40\% Recall with CoT on TCM samples, likely reflecting limited exposure to the philosophical and linguistic constructs that characterize TCM literature. In contrast, LLaMA3.1-8B and Baichuan2-13B-Chat tend to maintain robust Recall but occasionally sacrifice Precision, leading to moderate F1 scores. This suggests that while they are adept at identifying a broad range of metaphorical patterns, they also over-detect patterns that do not necessarily align with TCM theory, generating moderate amounts of false positives.

A closer qualitative examination reveals that the most frequently misread metaphors involve subtle doctrinal concepts, such as “Yang collapsing,” “Obstruction by blood stasis,” or compound pathologies combining heat, dampness, and deficiency. In multiple trials, models without CoT reasoning tended to interpret these symbolic TCM descriptions as either nonsensical or superficially mapped to Western inflammatory processes. In one example, “Yang collapsing” was incorrectly labeled by some single-agent baselines as “systemic autoimmune response,” a mismatch that arose because the literal mention of collapse was conflated with the notion of immune dysregulation. The multi-agent CoT approach mitigates such confusion by allowing a specialized TCM agent to parse the deeper cultural and clinical significance of “Yang collapsing” before handing off to the WM expert agent, which then aligns any physiologically relevant aspects (e.g., hypotension or shock symptoms). Furthermore, when asked to confirm domain plausibility, the Coordinator agent often flags incongruities in the bridging process, reducing the probability of spurious or “hallucinated” alignments.

Ablation studies confirm the importance of both the specialized agents and explicit CoT prompting. Removing the TCM Expert agent, for instance, reduces Accuracy by nearly 10\% on TCM metaphors, underscoring the need for domain-tailored parsing. Similarly, excluding the WM Expert agent typically impairs the system’s ability to anchor metaphorical inferences to pathophysiological correlates, causing a spike in false positives where TCM terms were erroneously accepted as valid WM diagnoses. Most critically, eliminating CoT outputs drastically degrades the system’s interpretability and leads to a 5--15\% drop in F1 on average, as seen by comparing Tables II--III and Tables IV--V. These findings highlight the central role played by stepwise reasoning in uncovering and correcting domain-level discrepancies between TCM and WM paradigms.

\begin{figure}[htbp]
\includegraphics[width=0.85\columnwidth]{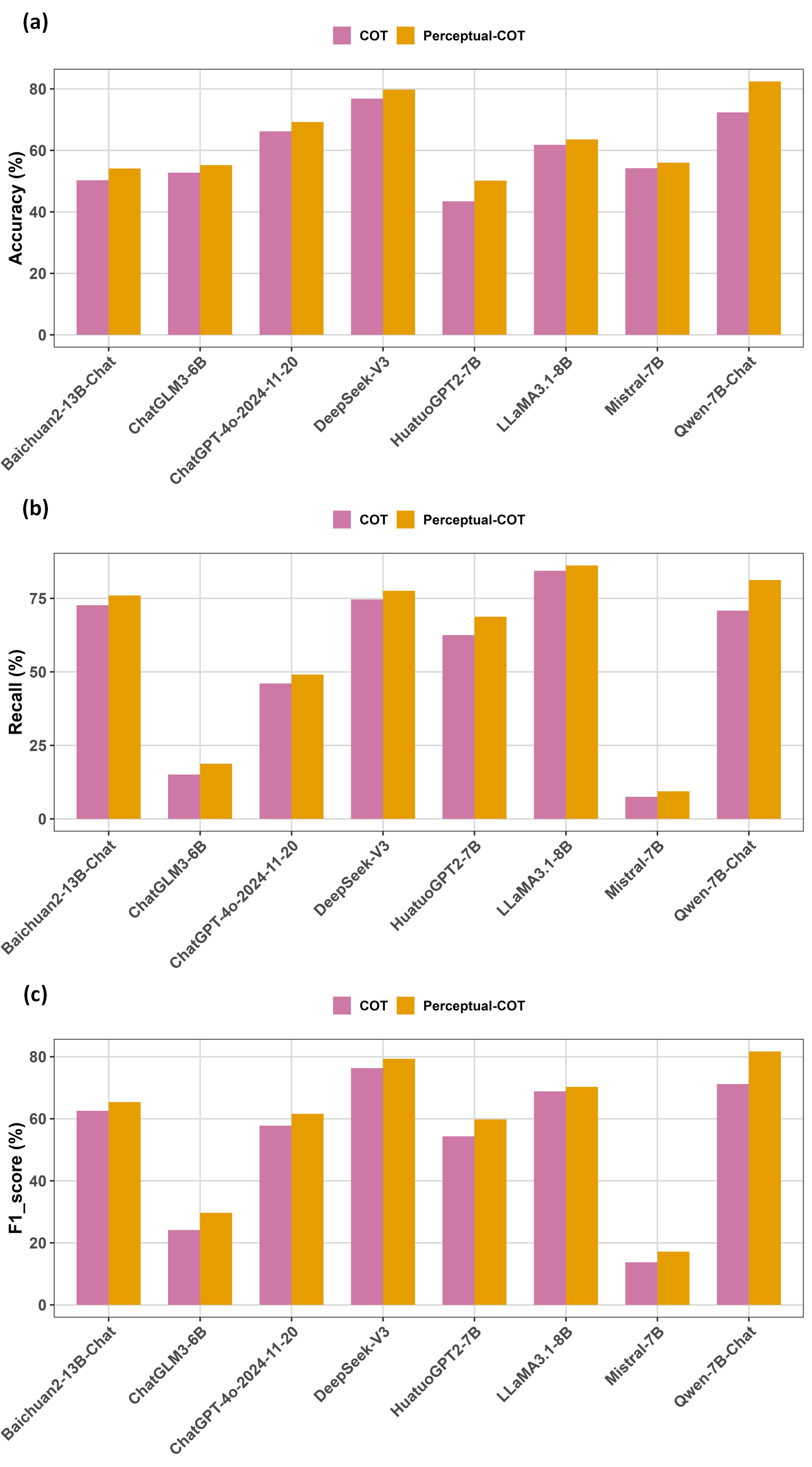}
\caption{TCM Result Visualization: COT vs Perceptual-COT}
\label{fig}
\end{figure}

\begin{figure}[htbp]
\includegraphics[width=0.85\columnwidth]{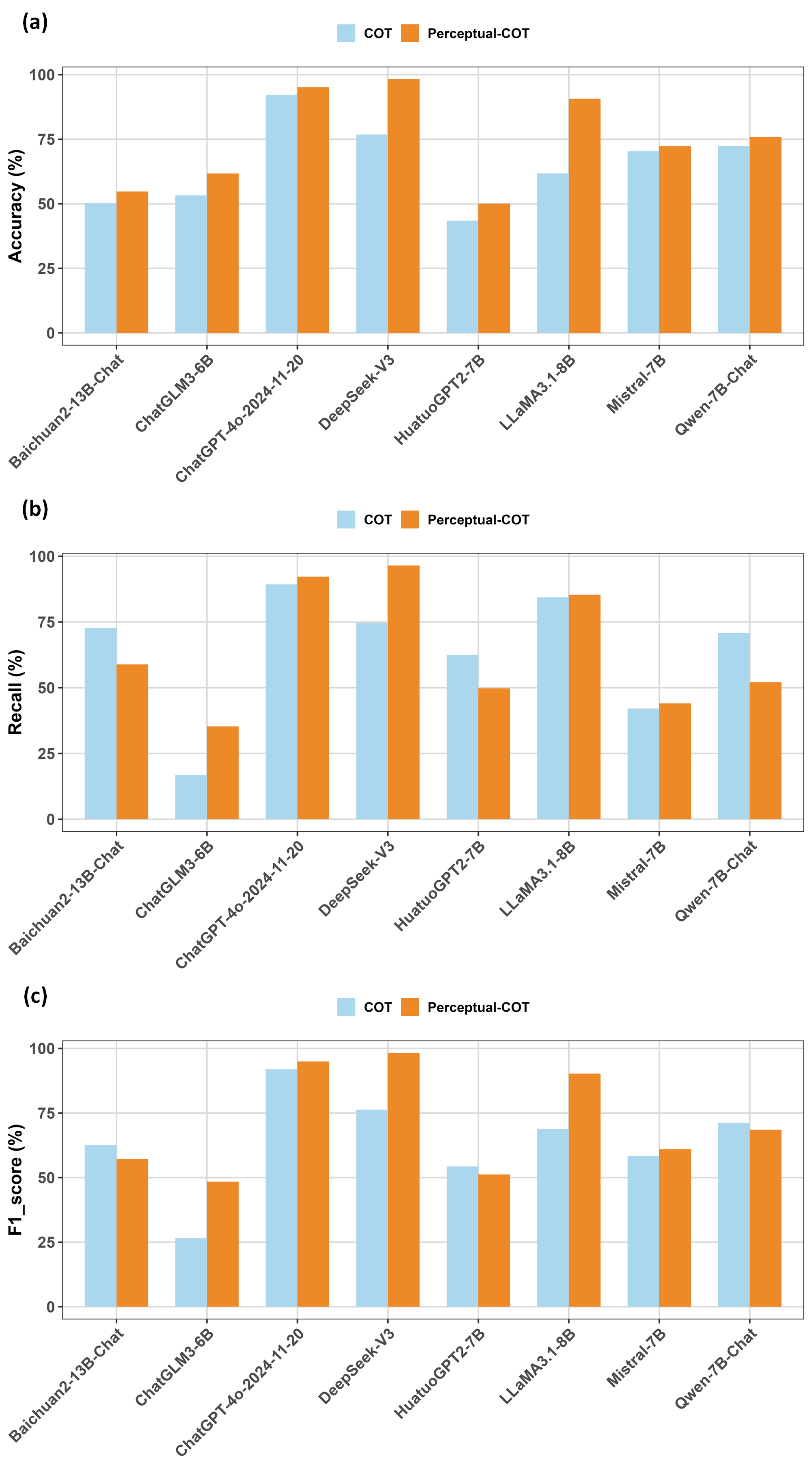}
\caption{WM Result Visualization: COT vs Perceptual-COT}
\label{fig}
\end{figure}

\section{Conclusion}
In this paper, we introduced an enhanced multi-agent and chain-of-thought framework that targets the complexities of interpreting TCM metaphors within a WM-oriented context. By allocating specialized roles to TCM and WM agents and coordinating through a meta-level oversight, our method alleviates conceptual disconnects and augments the interpretive precision of AI systems. Chain-of-thought prompting further exposes each reasoning path, granting domain experts a transparent audit of how final outputs are derived.
From a broader standpoint, we underscore that TCM metaphors are foundational elements encoding profound cultural, philosophical, and clinical knowledge. Our approach respects these conceptual intricacies by mapping them to WM correlates with minimal distortion, opening up a richer dialogue across medical paradigms. Beyond clinical decision support, our findings may shape medical education and encourage further advancements in integrative medicine.
Moving forward, we aim to validate our architecture with extensive empirical studies, encompassing multilingual data, additional TCM schools, and more nuanced metaphorical forms spanning emotional and spiritual dimensions. We believe this multi-agent, chain-of-thought approach can become a cornerstone for reconciling heterogeneous medical traditions, enabling deeper synergy between TCM and WM and contributing to improved global health outcomes.


\bibliographystyle{IEEEtran}


\end{CJK}

\end{document}